\documentclass[pdflatex,sn-mathphys-num]{sn-jnl}

\usepackage{subfigure}
\usepackage{graphicx}%
\usepackage{multirow}%
\usepackage{amsmath,amssymb,amsfonts}%
\usepackage{amsthm}%
\usepackage{mathrsfs}%
\usepackage[title]{appendix}%
\usepackage{xcolor}%
\usepackage{textcomp}%
\usepackage{manyfoot}%
\usepackage{booktabs}%
\usepackage{algorithm}%
\usepackage{algorithmicx}%
\usepackage{algpseudocode}%
\usepackage{listings}%

\theoremstyle{thmstyleone}%
\theoremstyle{thmstyletwo}%

\theoremstyle{thmstylethree}%

\begin{document}

\title[Article Title]{Scaling Particle Collision Data Analysis}

\author*[1]{\fnm{Hengkui} \sur{Wu}} \email{hkwu@ssymmetry.com}
\equalcont{These authors contributed equally to this work.}
\author[1]{\fnm{Panpan} \sur{Chi}} \email{ppchi@ssymmetry.com}
\equalcont{These authors contributed equally to this work.}
\author[2]{\fnm{Yongfeng} \sur{Zhu}} \email{zhuyongfeng@pku.edu.cn} 
\author[1]{\fnm{Liujiang} \sur{Liu}} \email{liuliujiang@ssymmetry.com}
\author[1]{\fnm{Shuyang} \sur{Hu}} \email{syhu@ssymmetry.com}
\author[3,5]{\fnm{Yuexin} \sur{Wang}} \email{wangyuexin@ihep.ac.cn}
\author[2]{\fnm{Chen} \sur{Zhou}}
\author[1]{\fnm{Qihao} \sur{Wang}} \email{qhwang@ssymmetry.com}
\author[1]{\fnm{Yingsi} \sur{Xin}} \email{ysxin@ssymmetry.com}
\author[1]{\fnm{Bruce} \sur{Liu}} \email{bliu@ssymmetry.com}
\author[1]{\fnm{Hengyuan} \sur{Wu}} \email{hy@ssymmetry.com}
\author[1]{\fnm{Dahao} \sur{Liang}}
\author[1]{\fnm{Xinglong} \sur{Jia}}

\author*[3,4]{\fnm{Manqi}\sur{Ruan}}\email{ruanmq@ihep.ac.cn}

\affil*[1]{\orgname{Supersymmetry Technologies}, \orgaddress{\street{1528 Gumei Road,Xuhui District}, \city{Shanghai}, \postcode{200000}, \country{China}}}
\affil[2]{\orgdiv{State Key Laboratory of Nuclear Physics and Technology}, \orgname{ School of Physics, Peking University}, \orgaddress{ \city{Beijing}, \postcode{100871}, \country{China}}}
\affil*[3]{\orgdiv{Institute of High Energy Physics}, \orgname{Chinese Academy of Science}, \orgaddress{\street{ 19B Yuquan Road, Shijingshan District}, \city{Beijing}, \postcode{10049}, \country{China}}}
\affil*[4]{\orgdiv{University of Chinese Academy of Sciences}, \orgname{Chinese Academy of Science}, \orgaddress{\street{  19A Yuquan Road, Shijingshan District}, \city{Beijing}, \postcode{10049}, \country{China}}}

\affil[5]{\orgdiv{China Center of Advanced Science and Technology}, \orgaddress{\city{Beijing}, \postcode{100190}, \country{China}}}

\abstract{

For decades, researchers have developed task-specific models to address scientific challenges across diverse disciplines. Recently, large language models (LLMs) have shown enormous capabilities in handling general tasks; however, these models encounter difficulties in addressing real-world scientific problems, particularly in domains involving large-scale numerical data analysis, such as experimental high energy physics. This limitation is primarily due to BPE tokenization's inefficacy with numerical data. In this paper, we propose a task-agnostic architecture, BBT-Neutron, which employs a binary tokenization method to facilitate pretraining on a mixture of textual and large-scale numerical experimental data.  We demonstrate the application of BBT-Neutron to Jet Origin Identification (JoI), a critical categorization challenge in high-energy physics that distinguishes jets originating from various quarks or gluons. Our results indicate that BBT-Neutron achieves comparable performance to state-of-the-art task-specific JoI models. Furthermore, we examine the scaling behavior of BBT-Neutron's performance with increasing data volume, suggesting the potential for BBT-Neutron to serve as a foundational model for particle physics data analysis, with possible extensions to a broad spectrum of scientific computing applications for Big Science experiments, industrial manufacturing and spacial computing. The project code is available at \url{https://github.com/supersymmetry-technologies/bbt-neutron}.

}




\maketitle

\section{Introduction}\label{sec1}


Scaling law plays a critical role in developing emergent abilities and in-context learning of large language models \cite{Gage1994ANA,bommasani2021on,wei2022emergent,kaplan2020scaling}. 
Pretraining LLM on a sufficiently broad and heterogeneous corpus gives rise to in-context learning building upon the processes of absorbing skills within the repetitive sequences of subsets of data which usually come as natural language  \cite{radford2019language,brown2020language}. 
The efficacy of in-context learning is directly proportional to the volume and variety of the data, as well as the computational power expended during the pretraining phase, adhering to a power-law distribution \cite{hoffmann2022training}. 
The ``emergence'' of LLM, described by the ability to solve specific tasks jumping up abruptly when data and model size pass a certain threshold, is attributed to the scaling of pretraining.   

Scaling unifies the pretraining processes for vision, voice, and text within a common framework, contributing to the success of recent multimodal models, including Sora \cite{liu2024sora}, GPT-4V \cite{yang2023gpt}, image modeling \cite{zhai2022scalingvisiontransformers,peebles2023scalablediffusionmodelstransformers},video modeling \cite{henighan2020scalinglawsautoregressivegenerative} and others \cite{liu2024neuralscalinglawsgraphs,hilton2023scalinglawssingleagentreinforcement}. However, foundational models encounter significant challenges when attempting to utilize large-scale scientific experimental data. A substantial amount of data generated in scientific research, industrial manufacturing, and space computing is stored in numerical formats. The numerical representations derived from spatial and temporal measurements constitute a crucial layer of abstraction of the physical world. These representations are essential for constructing world models, which have been the focus of extensive research as a pathway toward achieving artificial general intelligence \cite{gupta2021embodied,garrido2024learning}. General LLMs face notable challenges in handling experimental science tasks with high numerical data concentration, primarily due to the inherent difficulties in the following aspects.

First, the mainstream number representation method, Byte Pair Encoding (BPE) tokenization, as adopted by GPT-3 \cite{radford2019language,brown2020language}, limits large language models’ (LLMs) performance in numerical processing and reasoning tasks. BPE tokenization, a simple data compression technique that iteratively replaces the most frequent byte pairs with a single, unused byte can introduce ambiguity when tokenizing numbers, as it often fragments numerical values into arbitrary segments, thereby obscuring their original meaning. Additionally, BPE’s statistical approach may result in inconsistent tokenization of identical numbers across different contexts, further complicating downstream numerical tasks.
Llama3 \cite{dubey2024llama} attempts to address this by splitting numbers into indi-vidual tokens to improve consistency in number representation. However, even the most advanced number tokenization techniques cannot eliminate the fundamental issue: the quantitative relationships between individual numbers are distorted post-tokenization, impairing foundational models’ ability to perform basic comparisons and arithmetic operations \cite{yang2023gpt,yuan2023well,McLeish2024TransformersCD}.
This limitation has driven consensus on the necessity of advancing number tokenization methods to enable significant improvements in LLM applications for scientific tasks. 

Second, task-specific specialist models have been the primary drivers of AI advancements in scientific domains over recent decades \cite{abramson2024accurate,jumper2021highly}. Researchers have leveraged domain-specific structures to develop customized supervised learning models for scientific tasks, which typically require data that are identically and independently distributed (I.I.D). However, these task-specific models lack the advantages of transfer learning across multiple tasks and often perform poorly in out-of-distribution generalizations—scenarios inherently conducive to scientific discovery. Increasingly complex structures in specialist models have not yielded performance improvements comparable to task-agnostic generalist models, which benefit from scaled data and computing \cite{Chung}.

Third, foundational models have been applied to scientific tasks in areas such as question answering \cite{taylor2022galactica,xie2023darwin}, quantum physics calculations \cite{pan2024quantum}, prime number distribution \cite{romera2024mathematical}, partial differential equation (PDE) computation \cite{shen2024ups}, and time series regression \cite{vacareanu2024words}. However, to date, LLMs developed for scientific applications are primarily trained on symbolic data, including text and scientific symbols such as DNA sequences and mathematical formulas. Numerical experimental data particularly from large-scale scientific projects in fields like particle physics and astronomy are typically excluded from these training corpora due to the absence of a general architecture capable of multimodal synthesis of text and experimental data. This limitation significantly hinders the capacity of general LLMs to address complex, real-world scientific problems.

A new paradigm for task-agnostic architecture in scientific computing is therefore required to enable in-context learning and to uncover scaling laws applicable to experimental data analysis. To address this requirement, we have developed Big Bang Transformer-Neutron (BBT-Neutron), a task-agnostic large language model architecture specifically designed for scientific tasks, featuring an innovative number tokenization approach called Binary Tokenization. This binary-native approach encodes input data as byte sequences, preserving the intrinsic structure and quantitative integrity of numerical data and avoiding ambiguities caused by segmenting or merging numeric and textual information. Binary Tokenization demonstrates robust capabilities in unifying representation of diverse data modalities, including text, numerical values, images. A significant proportion of scientific data generated in large-scale experiments is stored in binary formats. This characteristic enables direct input into the Binary Tokenization framework without requiring additional preprocessing steps, thereby streamlining the process and facilitating fully end-to-end model training. The performance of BBT-Neutron is then benchmarked with the task of JoI, a challenge and critical classification algorithm for the experimental particle physics using high energy collider.

\begin{figure}[h]
    \centering
    \includegraphics[width=.6\textwidth]{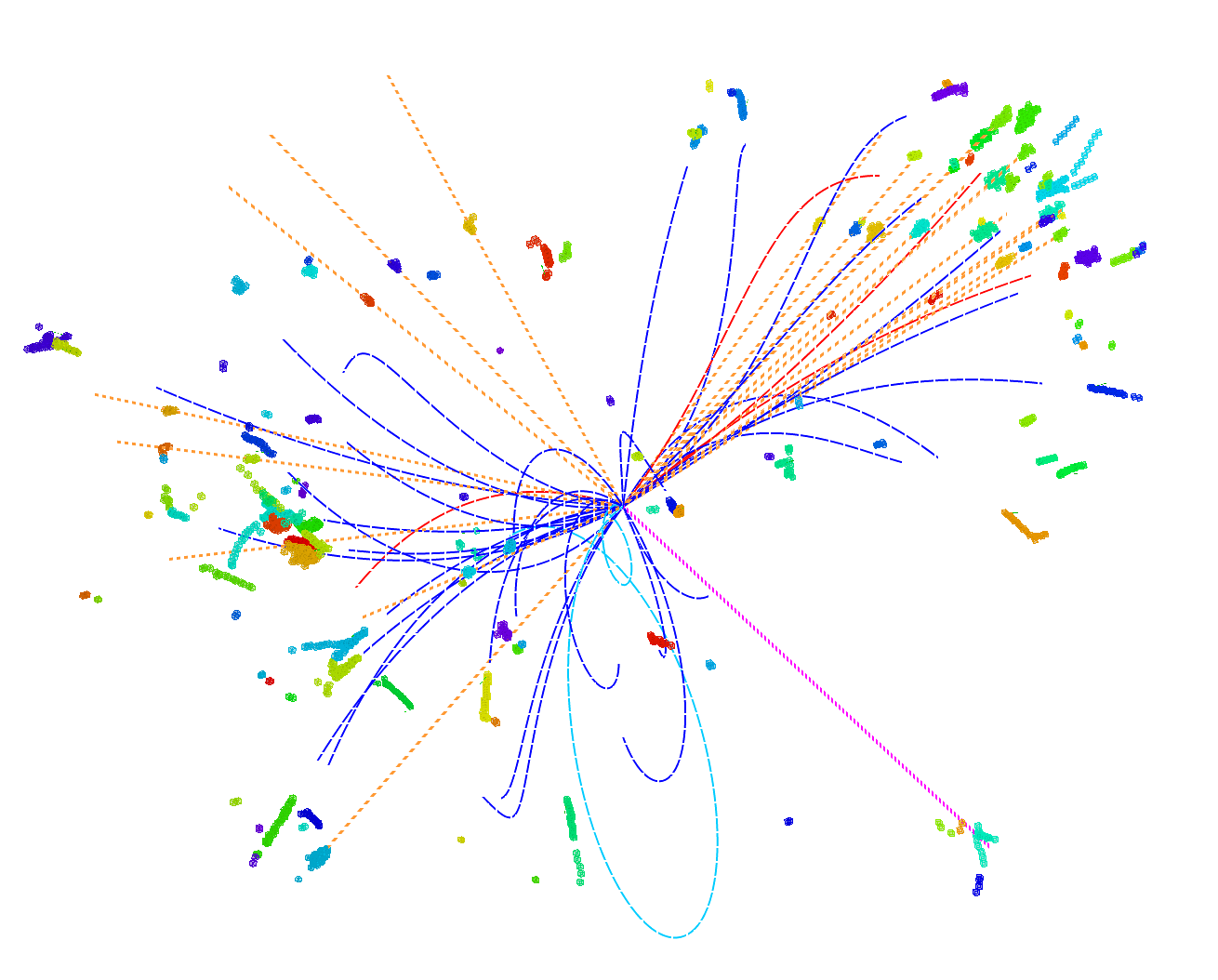}
    \caption{ Event display of an $e^+e^-\rightarrow \nu\bar{\nu} H\rightarrow \nu\bar{\nu} gg$ ($\sqrt{s}$ = 240 GeV) event simulated and reconstructed with the CEPC baseline detector~\cite{CEPC_CDR_Phy}.
    Different particles are depicted with colored curves and straight lines: \textcolor{red}{red} for $e^{\pm}$, \textcolor{cyan}{cyan} for $\mu^{\pm}$, \textcolor{blue}{blue} for $\pi^{\pm}$, \textcolor{orange}{orange} for photons, and \textcolor{magenta}{magenta} for neutral hadrons.
\label{fig:Display}}
\end{figure}

JoI is a newly developed idea for experimental particle physics that aims to distinguish jets originating from different quarks and gluons, which are the building blocks of our material world and are fundamental particles in the Standard Model of Particle Physics. 
Once quark or gluon is produced in high-energy collisions, such as the Large Hadron Collider (LHC) and future electron-positron Higgs factories, it would immediately give rise to a spray of particles, mostly hadrons, that travel in the same direction.
This spray of particles is conventionally known as a jet, which is a critical object for physics measurements in collider experiments.
Significant efforts have been devoted to identifying the origin of jets, specifically, determining which type of colored particle initiated the jet. 
Identification of jet origin is central to many physics analyses, particularly in studies of the Higgs, W, and Z bosons, as nearly 70\% of these bosons decay into two jets directly.
In addition, jets are fundamental to advancing our understanding of QCD.  
However, jets originating from different types of colored particles exhibit only marginal differences in their observables, making accurately identifying the origin of the jet extremely challenging.

JoI was incarnated in the specific algorithms of jet flavor tagging, jet charge measurement, and strange, light, and gluon jet tagging. 
At the Large Electron-Positron Collider (LEP), the predecessor to the LHC, jet origin identification initially relied on cuts applied to various discriminating variables to distinguish between b-, c-, and light-jets \cite{aleph2000study,abe1998measurement}. 
Over time, likelihood-based methods \cite{borisov1998combined,abdallah2003b} emerged, combining multiple observables into a single discriminant. 
With advancements in detector technology, the LHC provided improved reconstruction performance, coupled with advanced algorithms such as Boosted Decision Trees (BDTs) \cite{cms2018identification,cms2016identification}, followed by recurrent neural networks (RNNs) \cite{egan2017long,guest2016jet}, convolutional neural networks (CNNs) \cite{cogan2015jet,de2016jet}, Graph Neural Network (GNN), and transformer, the origin of jet has been identified with better performance.
JoI was then developed using GNN-based ParticleNet and Transfomer-based Particle Transformer; it could simultaneously identify 11 jet species: five quarks, five anti-quarks, and gluons at the proposed electron-positron Higgs factory. 
JoI offers an unprecedented distinguishing power of different jet species and is critical for almost all the physics measurements with the jet's final state. 
Quantitative analysis shows that the anticipated accuracies of relevant Higgs measurements could be improved from two folds up to two orders of magnitude using JoI, and thus significantly boost the discovery power of collider experiments.


We observe that BBT-Neutron achieves a performance comparable to that of specialized models like ParticleNet and Particle Transformer. As the first task-agnostic general LLM architecture capable of pretraining on a multimodal mixture of textual and large-scale numerical datasets, BBT-Neutron demonstrates state of the art performance in JoI, compared to the results of leading domain-specific models. This achievement marks a significant step toward developing foundational models for scientific research applications that can generalize across diverse data types and tasks.

\section{Results}
\label{result}

Jason et al.\cite{wei2022emergentabilitieslargelanguage} investigated the emergent abilities of large language models by examining the scaling behavior of model performance relative to model size and compute resources, while excluding dataset size due to specific experimental constraints. More recently, Fanqi et al.\cite{lin2024datascalinglawsimitation} explored the data scaling laws in robotics applications.   Despite these advancements, further research is required to comprehensively understand the scaling phenomena in numerically intensive models, particularly in the context of experimental scientific domains. 
\begin{figure}[h]`
    \centering
    \includegraphics[width=.45\textwidth]{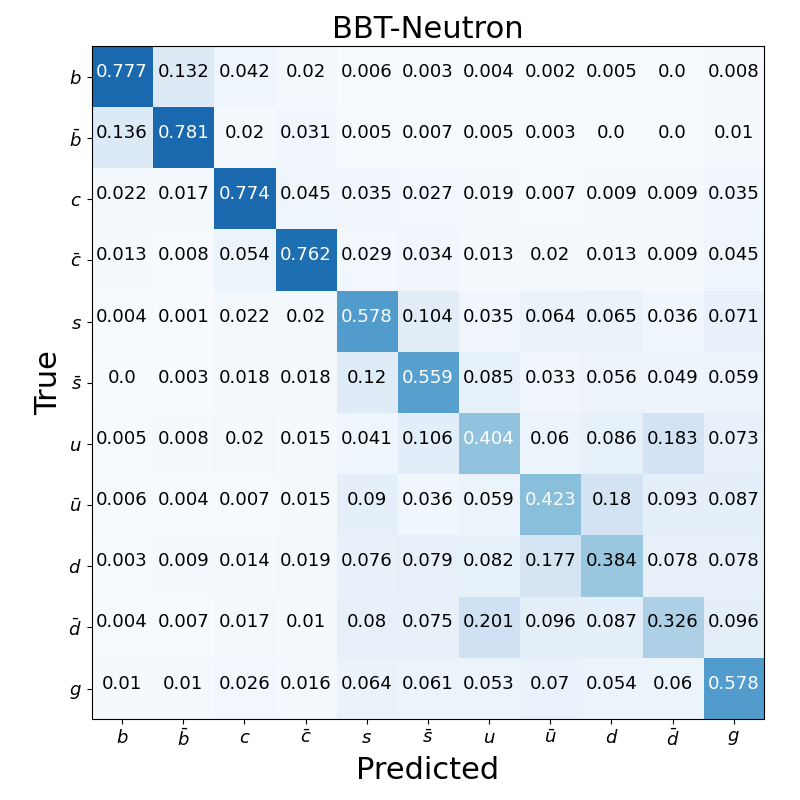}
    \\
    \includegraphics[width=.45\textwidth]{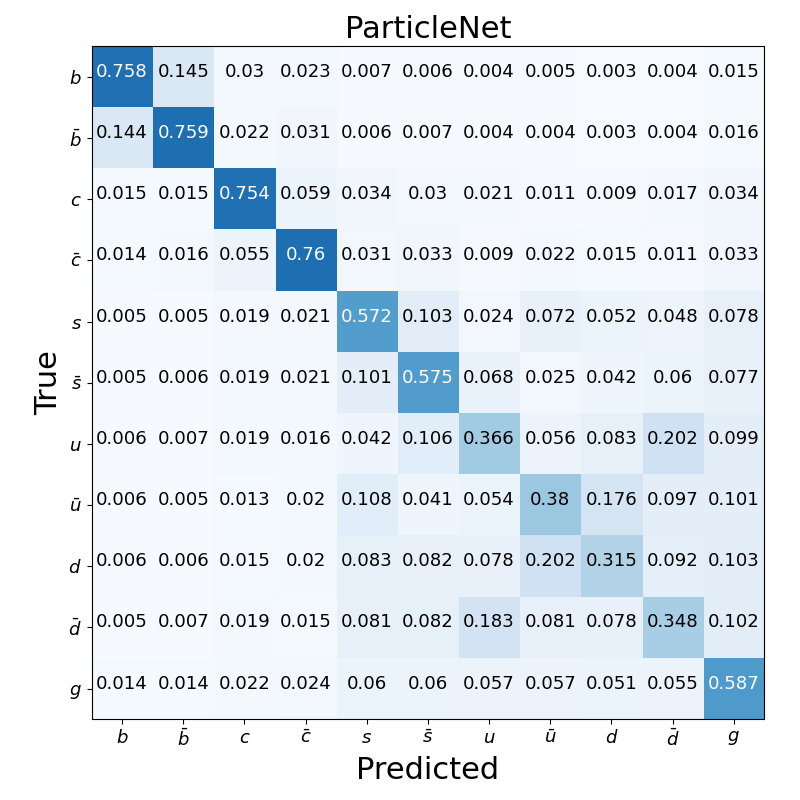}
    \hfill
    \includegraphics[width=.45\textwidth]{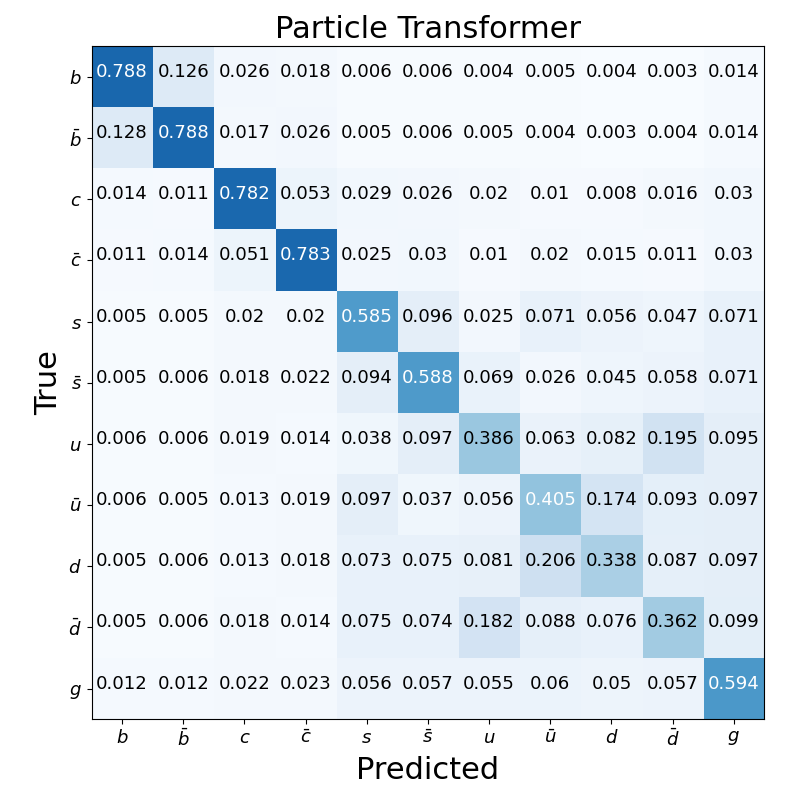}
    \caption{ With the statistics of each jet one million, 60\% of them used for training, 20\% for validation, and another 20\% for testing, the confusion matrix $M_{11}$ obtained by BBT-Neutron, ParticleNet, and Particle Transformer for $\nu\bar{\nu}H, H\to jj$ events at 240 GeV center-of-mass energy. Each matrix is normalized to unity for each truth label (row).
\label{fig:M11}}
\end{figure}
\begin{figure}[h]
    \centering
    \includegraphics[width=.5\textwidth]{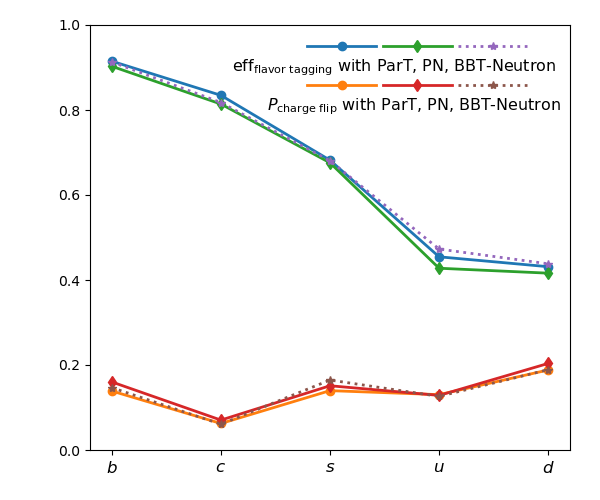}
    \caption{ Jet flavor tagging efficiencies and charge flip rates for each quark species with ParticleNet and Particle Transformer.
\label{fig:effFlip}}
\end{figure}
 

We compare the performance of BBT-Neutron and two state-of-the-art models, ParticleNet(PN) \cite{Qu:2019gqs} and Particle Transformer(ParT) \cite{Qu:2022mxj}. The results for these models are shown in  Fig~\ref{fig:M11},  using an 11-dimensional confusion matrix  $M_{11}$, which classifies each jet according to the category with the highest predicted score. In the quark sector, the $M_{11}$ matrix is approximately symmetric and can be block-diagonalized into $2 \times 2$ blocks, each corresponding to a specific quark species. This confusion matrix provides a comprehensive overview of the model’s classification performance, highlighting both correct and incorrect predictions across various jet categories. The top panel of Fig~\ref{fig:M11} presents the overall JoI performance for BBT-Neutron.
These two state-of-the-art deep learning models show approximately similar performance as that of the BBT-Neutron.
To further assess JoI performance, two key metrics are analyzed: jet flavor tagging efficiencies and charge flip rates. 
Jet flavor tagging efficiency is defined as half the sum of the values within each block, which does not distinguish the jet originating from quark and anti-quark.
The charge flip rate is the ratio of the off-diagonal elements to the total sum of the block, which represents the probability of misidentification of the jet originating from quark and anti-quark. 
Figure~\ref{fig:effFlip} presents the flavor tagging efficiency and charge flip rates for each quark species, demonstrating that PN, ParT, and BBT-Neutron exhibit comparable performance at 10 million statistic.
\begin{figure}[htbp]
\centering
        \subfigure[]{ \label{fig:PhotonAngle}
		\includegraphics[width=0.3\textwidth]{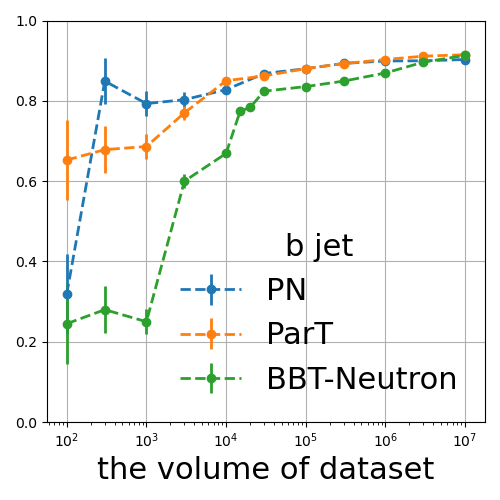}}
	\subfigure[]{ \label{fig:PhotonSeparation}
		\includegraphics[width=0.3\textwidth]{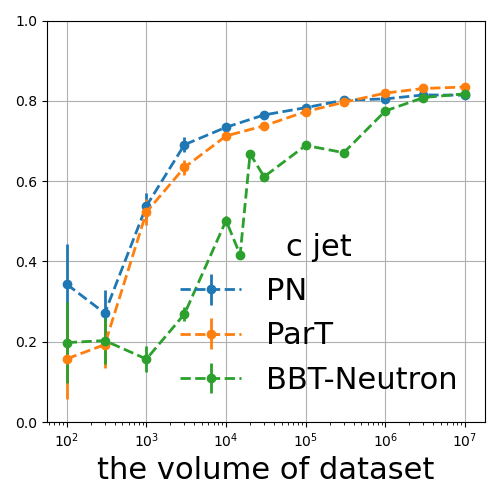}}
  \\
        \subfigure[]{ \label{fig:dedx2rho}
		\includegraphics[width=0.3\textwidth]{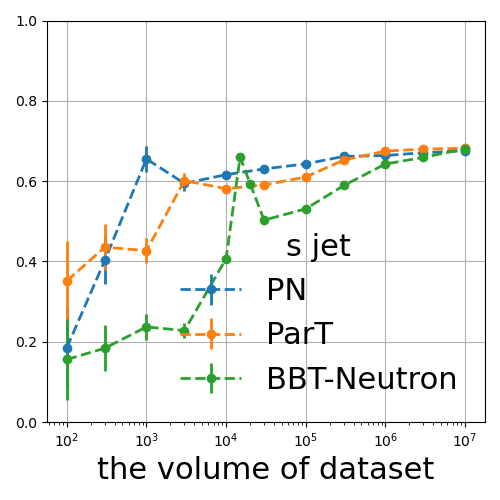}}
  	\subfigure[]{ \label{fig:dedx2bg}
		\includegraphics[width=0.3\textwidth]{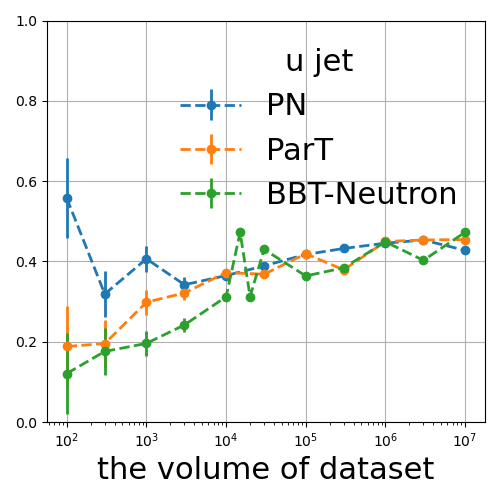}}
  	\subfigure[]{ \label{fig:dedx2Cos}
		\includegraphics[width=0.3\textwidth]{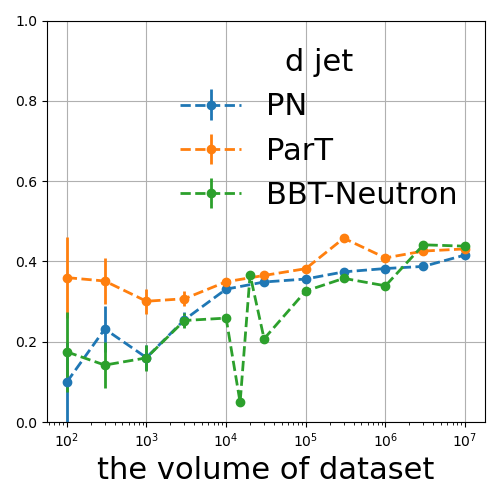}}
	\caption{ \label{fig:eff} The scaling behavior of BBT-Neutron, ParticleNet (PN), and Particle Transformer (ParT) in terms of jet flavor tagging efficiency as a function of training data volume is illustrated. Panels (a), (b), (c), (d), and (e) correspond to the bottom, charm, strange, up, and down quark flavors, respectively. }
\end{figure}

\begin{figure}[htbp]
\centering
\subfigure[]{ \label{fig:PhotonAngle}
		\includegraphics[width=0.3\textwidth]{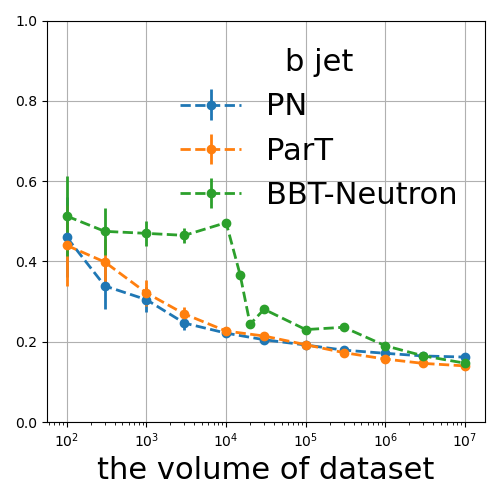}}
	\subfigure[]{ \label{fig:PhotonSeparation}
		\includegraphics[width=0.3\textwidth]{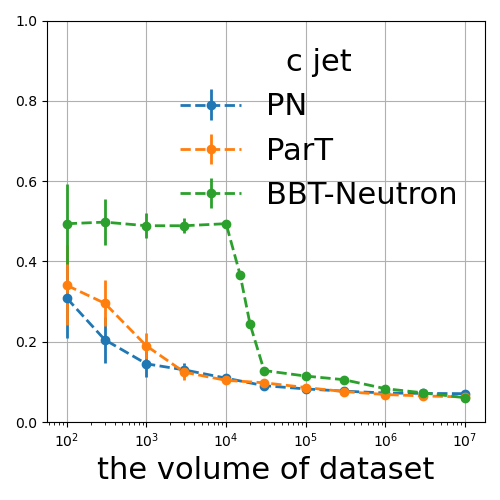}}
  \\
        \subfigure[]{ \label{fig:dedx2rho}
		\includegraphics[width=0.3\textwidth]{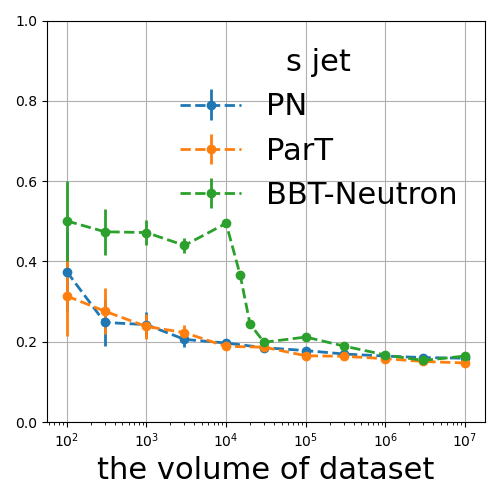}}
  	\subfigure[]{ \label{fig:dedx2bg}
		\includegraphics[width=0.3\textwidth]{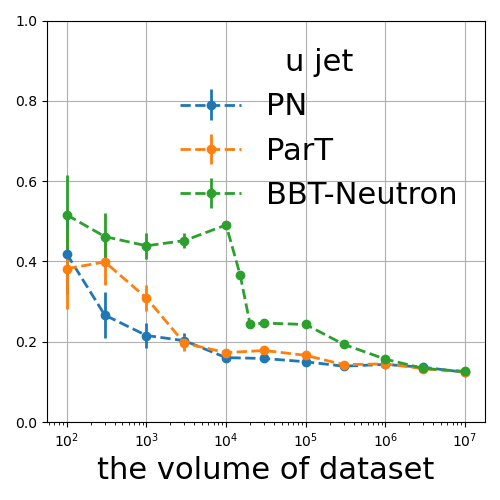}}
  	\subfigure[]{ \label{fig:dedx2Cos}
		\includegraphics[width=0.3\textwidth]{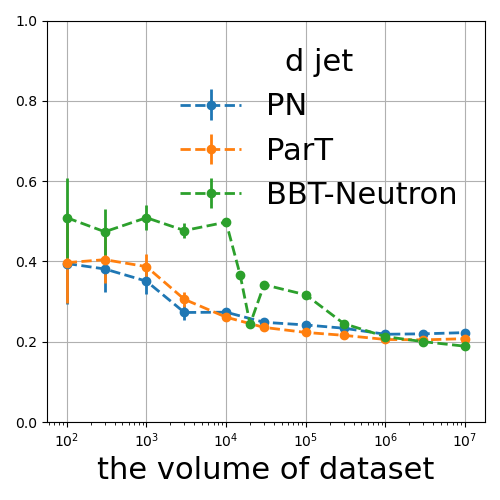}}
	\caption{ \label{fig:rate} The scaling behavior of BBT-Neutron, ParticleNet (PN), and Particle Transformer (ParT) in terms of jet charge flip rate as a function of training data volume is illustrated. Panels (a), (b), (c), (d), and (e) correspond to the bottom, charm, strange, up, and down quark flavors, respectively. }
\end{figure}
 
 We conduct a comparative analysis on the scaling behaviors of BBT-Neutron and two state-of-the-art models, PN and ParT, applied to JOI across training datasets ranging from 100 to 10 million events. All three models show significant performance increase to dataset size scaling, forming an S-curve pattern in both flavor tagging efficiency and jet charge flip rates, as shown in Figures~\ref{fig:eff} and~\ref{fig:rate}, respectively. Notably, distinct scaling behaviors were observed between BBT-Neutron and the specialized models. 
The critical data thresholds on the S-curves for flavor tagging and charge flips indicate distinct emergence behaviors in BBT-Neutron. These behaviors are not as pronounced in the specialized models.
In flavor tagging, PN and ParT, both specialized models, exhibit similar performance and scaling trends relative to the training data. This similarity is due to key structural and operational principles shared by both models, as described in Methods section~\ref{PNParT}. At smaller dataset sizes, these specialized models produce meaningful results, whereas BBT-Neutron initially performs at levels akin to random guessing. As the dataset size increases, all models exhibit performance gains, with the specialized models showing smooth improvement before reaching their respective performance plateaus. In contrast, BBT-Neutron requires larger datasets to achieve similar levels of performance, leading to a sharper jump on the curve. Particularly for b, c, and s quarks, BBT-Neutron requires approximately an order of magnitude more data than specialized models to achieve comparable performance. Once BBT-Neutron surpasses a critical dataset size threshold, its performance converges with that of the specialized models. For b,s and u quarks, BBT-Neutron exhibits a trend of surpassing the performance of the specialized models with further training on larger datasets.

 In the task of jet charge measurement a binary classification problem aimed at distinguishing jets originating from quarks of the same flavor but opposite charge. BBT-Neutron initially demonstrates near-random performance, followed by a sharp improvement to above-random levels. This behavior is reminiscent of the emergent capabilities observed in large language models \cite{wei2022emergentabilitieslargelanguage}. Specifically, BBT-Neutron produces random predictions when trained on datasets containing fewer than 10,000 events. However, its performance improves markedly beyond this threshold, suggesting that the model begins to capture complex, high-dimensional features after surpassing a critical data size. For datasets exceeding 30,000 events, BBT-Neutron's scaling behavior aligns more closely with that of specialized models, exhibiting smoother performance gains with further increases in data size. By the time the dataset reaches approximately 3,000,000 events, BBT-Neutron achieves jet charge flip rates comparable to those of domain-specific models, which display steady performance improvements across all dataset sizes.
The observed sharp jump in performance, from near-random to above-random levels, with increasing training dataset size in both flavor tagging and charge flip tasks for BBT-Neutron, strongly indicates the emergence of a generalist architecture. A plausible explanation for the absence of such emergent behaviors in specialist architectures, such as PN and ParT, is the incorporation of domain-specific structural biases. These models employ specialized architectural designs tailored to represent particle interactions and classification, which may lead to faster saturation of performance as data scales. In contrast, the generalist BBT-Neutron architecture processes all physical structures uniformly using a common data representation. Unlike specialist architectures, which leverage permutative invariance-a key feature aligned with fundamental symmetry laws-by eliminating positional encodings or related operations, BBT-Neutron employs a left-to-right sequence input, consistent with the sequence-to-sequence paradigm of language models. While this approach requires larger datasets for the generalist architecture to infer equivalent structures, it enables a pronounced performance leap once a critical dataset threshold is surpassed. The results further suggest that BBT-Neutron is capable of learning spatial symmetries embedded in the data, even without explicitly incorporating permutative invariance as in specialized models.

\section{Discussion and Future Work}\label{notesummary}

Contemporary scientific research relies on vast experimental infrastructures, such as particle colliders in high-energy physics, that generate enormous volumes of data. A substantial portion of researchers’ efforts is focused on developing specialized models tailored for specific analytical tasks within these datasets. In contrast, task-agnostic architectures offer the potential to support diverse analytical tasks within a single foundational model, increasing efficiency and enabling performance gains through transfer learning and multi-dimensional data integration.

In this study, we apply BBT-Neutron with a task-agnostic architecture designed to address challenges in scientific problems enriched with large-scale numerical datasets to the complex task of Jet Origin Identification (JoI) in high-energy physics, demonstrating that it achieves performance comparable to specialized AI tools like PN and ParT. Although BBT-Neutron requires larger datasets to match the accuracy of these specialized models, it scales effectively and delivers competitive results. This work underscores the potential of large language models for scientific domains that demand precise numerical representation and processing. BBT-Neutron lays the groundwork for further exploration of LLM applications in high-energy physics and other fields requiring sophisticated numerical data handling. Its scalability supports task-agnostic meta-learning and positions it as a versatile tool for data-intensive scientific tasks across disciplines.

Moreover, akin to the continuous phase transitions observed in the capabilities of large language models with increasing model size and computational scale, it is expected that generalist architectures like BBT-Neutron will exhibit further performance enhancements with continued training and scaling.
In contrast to specialized architectures, LLMs like BBT-Neutron represent a paradigm shift in addressing scientific problems. BBT-Neutron exemplifies the potential of task-agnostic architectures capable of handling diverse scientific tasks such as classification, clustering, and regression through natural language inputs with minimal modifications to its output layer. This architectural strategy offers several key advantages: 
\begin{itemize}
\item Reduced Development Overhead: BBT-Neutron eliminates the need for task-specific architectural design and domain-specific feature engineering. 

\item Enhanced Knowledge Transfer: Pre-trained language models enable broader knowledge transfer compared to the limited transfer learning capabilities of specialized architectures. 

\item Improved Implementation Efficiency: A single task-agnostic model can address multiple tasks, reducing the need to maintain separate specialized models. 
\end{itemize}
In future work, we will expand BBT-Neutron’s pretraining to encompass a wider range of tasks related to particle collision experiments, aiming to develop a foundational model for high-energy physics data analysis. Additionally, we plan to pretrain BBT-Neutron on a combination of high-quality textual and experimental data, enabling simultaneous learning of physical theories and experimental methodologies.

\section{Methods}
\label{method}

\subsection{BBT-Neutron}

The BBT-Neutron architecture, as a multimodal binary model, adheres to a general transformer decoder-only framework but incorporates specialized components designed to understand text, scientific symbols, numbers, and images. These enhancements enable BBT-Neutron to effectively interpret and process content rich in numerical and visual data, making it especially well suited for applications involving big scientific facilities, such as particle accelerators, telescopes, and nuclear fusion reactors.

Compared to other byte-level models \cite{wu2024languagemodelsbytemodels,yu2023megabytepredictingmillionbytesequences,slagle2024spacebytedeletingtokenizationlarge}, our model is fully trained from scratch without relying on any pretrained models. Moreover, our model does not incorporate complex additional modules for modeling byte sequences within patches. Instead, it employs a more general causal attention mechanism to address both inter-patch and intra-patch sequence modeling simultaneously.

The critical architectural components of BBT-Neutron are described in Section \ref{architecture}, and the tokenization method of BBT-Neutron is described in Section \ref{tokenization},Furthermore, data preparation and training details is outlined in Section \ref{Training}.

\subsubsection{Model architecture}
\label{architecture}

\begin{figure}[htbp]
    \centering
    \includegraphics[width=.95\textwidth]{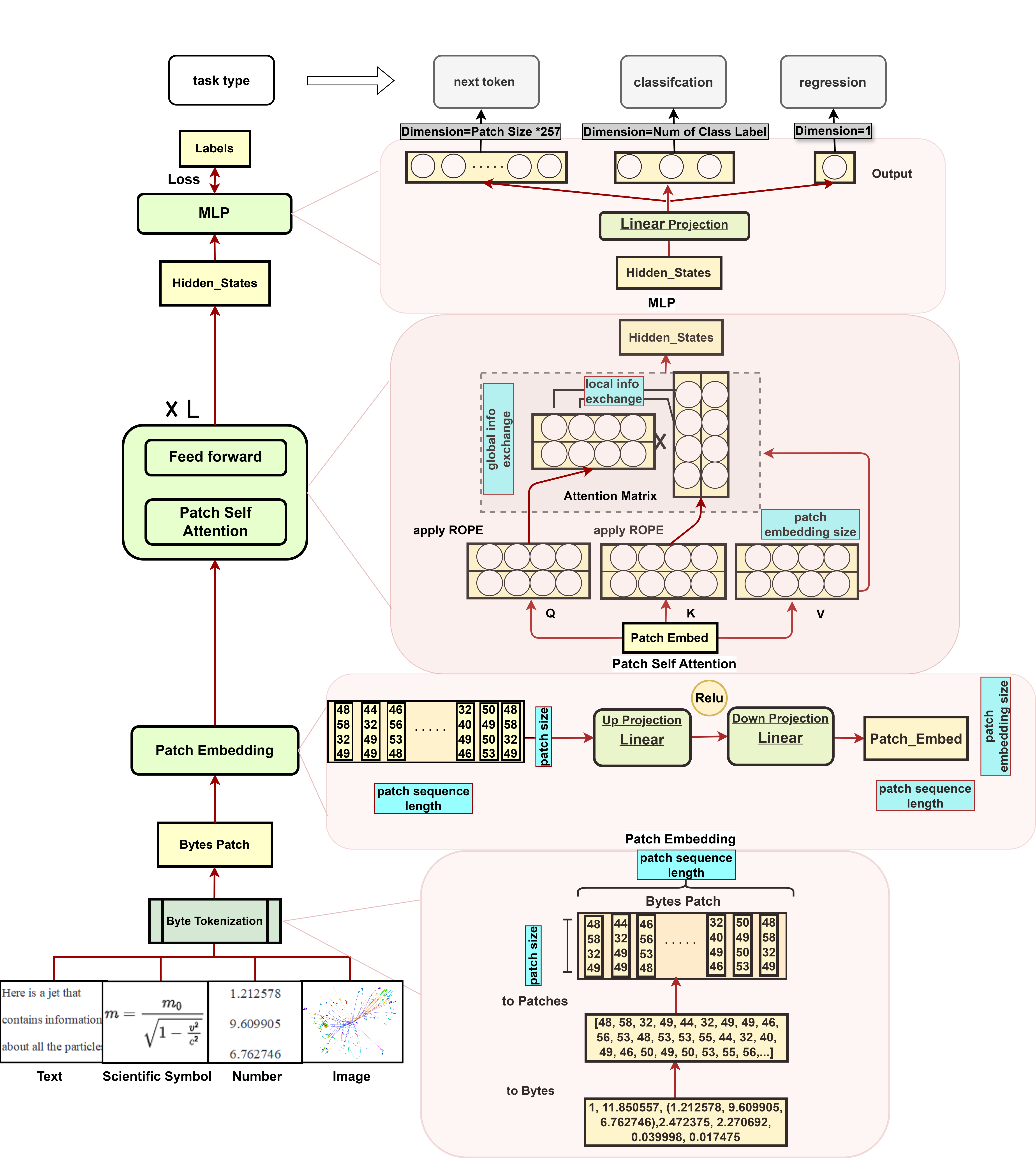}
    \caption{The overall architecture of BBT-Neutron that we propose, The green sections represent the model architecture modules, while the yellow sections indicate the inputs and outputs of these model modules.
\label{fig:BBT-Neutron_jet}}
\end{figure}

\textbf{Overview}
BBT-Neutron architecture involves converting input sequences into high-dimensional vectors through byte tokenization and patch embedding. The resulting embeddings pass through multiple transformer decoder layers. The model incorporates Rotary Position Embeddings (RoPE) \cite{SU2024127063} to infuse positional information into the self-attention mechanism, which allows our method to capture relative positional information between numbers, while Particle Transformer \cite{Qu:2022mxj} lacks position encoding. Each layer of the transformer decoder is endowed with causal self-attention capabilities, ensuring a sequential, left-to-right processing flow and allowing for the intricate interplay among tokens within patches. This mechanism allows the model to discern the subtleties of numerical relationships, thereby revealing the underlying physical laws. A feed-forward neural network enhances the model's ability to understand context and apply nonlinear transformations across layers. By layering these components, the model adeptly uncovers intricate dependencies within extended sequences. Finally, the output is transformed via a linear layer and a softmax function into a probability distribution over a vocabulary of size Patch Size$\times$257, predicting the next byte patch.

Furthermore, the BBT-Neutron model is versatile and can perform classification and regression tasks, which are common in many scientific applications where the goal is not necessarily to generate new sequences, but to classify inputs or predict continuous values \cite{ZHOU202057,Bannigan2023}. 
For classification tasks, the output dimension of the model corresponds to the number of distinct classes, allowing it to predict the most likely class label for a given input. In the case of regression tasks, the model outputs a single value representing the predicted outcome, such as a measurement or a score. This flexibility makes BBT-Neutron a powerful tool for a wide range of scientific analyses, from identifying patterns in experimental data to forecasting outcomes in complex systems.

BBT-Neutron includes two versions, patch and non-patch, which can be switched based on specific requirements.
\begin{itemize}
    \item Patch Version is a Patch-Based Model.
    It incorporates patches to enhance the overall computational speed of the model. As shown in Figure~\ref{fig:BBT-Neutron_jet},
    patches allow the model to process input sequences more efficiently by breaking them into smaller, manageable segments.
    The Patch Version model is designed to be computationally efficient, making it suitable for tasks where speed is a critical factor.
    \item Non-Patch Version is a  Byte-Level Model.
    It operates at the byte level, providing a more granular and detailed representation of the input data.
\end{itemize}

\textbf{Patch embedding} 
As illustrated in Figure~\ref{fig:BBT-Neutron_jet}, Patch Embedding involves two linear layers: the first projects the input patch into a higher-dimensional space, and the second refines this representation to produce the final embedding vector. Between these two linear layers, we introduce a ReLU activation function. This design choice enables the model to have non-linear expressive capabilities for the input byte patches, allowing it to capture more complex internal dynamics within the patches. Compared to similar byte-level models \cite{wu2024languagemodelsbytemodels}, which typically use a single linear layer for their embedding, our approach offers greater flexibility. This limitation in byte-level models restricts their ability to represent the intricate details and nonlinear interactions within the input patches.
For Patch Version, we designed an embedding layer that maps each byte patch to a dense vector representation. 
For Non-Patch Version, which is a byte-level version, we set the patch size to 1. This means that each byte is treated individually and that the embedding layer processes each byte separately. The embedding layer for the Non-Patch Version maps each individual byte to a dense vector representation. This approach provides a more granular and detailed representation of the input data, which can be crucial for tasks requiring fine-grained control and precision. 
This embedding layer is  tailored to capture the structural and semantic information of binary data.


\textbf{Patch Self-Attention}
The core of each Transformer block, which is also Attention Module as illustrated in Figure~\ref{fig:BBT-Neutron_jet}, is the multi-head self-attention mechanism. 
This mechanism allows the model to attend to different parts of the input sequence in parallel across multiple representation subspaces. 
First, the patch embed is linearly projected into query, key, and value vectors. 
The attention weights are then computed using the scaled dot-product attention mechanism, and the weighted value vectors are recombined to form the output. 
This enables the model to discern dependencies across various levels of the input sequence, bolstering its capacity to comprehend intricate semantics.
In the  patch self-attention mechanism, the attention operation is performed at the patch level.  Each patch embedding encapsulates information from all the bytes it comprises. Given that patch embeddings are vectorial, the matrix multiplication inherent in the attention mechanism not only enables the exchange of information across different patches but also fosters interaction among the bytes within individual patches. This two-tiered interaction empowers the model to effectively capture both local and global dependencies.

Patch Version models necessitate a careful approach to managing attention  masks within each decoder layer. Attention Mask is used to inform the model which parts of the input are actual data and which parts are padding, the model needs to disregard the padded parts because the padding tokens do not contain semantic information. The attention mask must be reshaped to correspond with shape of the patches. Following this, the boundary patches are marked as true for masking. This meticulous process ensures that the attention mechanism respects the patch boundaries accurately, thereby effectively identifying and integrating the inter dependencies both within individual patches and across different patches.

\textbf{LM Head}
The output dimension of the language model head layer in the patch-based model is defined as Patch Size $\times$ 257. 
Here, 257 represents the total number of byte IDs, which include the byte values from 0 to 255, plus the padding ID represented by 256, and Patch Size is the number of patches into which the text sequence is divided. 
This design allows the model to generate predictions for each patch independently, maintaining the efficiency and effectiveness of the patch-based approach.
In addition, for autoregressive generation in language models, generating byte-level tokens from patch-level representations presents certain challenges. Since patch-level embeddings summarize information from all bytes within a patch, they may lack some of the fine-grained details necessary for accurate byte-level token generation. This can make it more difficult to ensure that the generated tokens precisely reflect the intended content, especially in contexts where exact byte-level accuracy is crucial. To address this, the model may require additional mechanisms or fine-tuning to enhance the fidelity of byte-level token generation from patch-level embeddings. 

\subsubsection{Tokenization}
\label{tokenization}
Our tokenization method is engineered to accommodate various data modalities, encompassing text, scientific symbols, numerical data, and images. By using a binary tokenization approach, we can process all types of data uniformly, as they are stored in binary format in computers. Besides, to achieve a certain level of compression and improve computational efficiency, we employ a byte-based tokenization method, where each byte consists of 8 bits. This unified approach eliminates the need to develop specific tokenization methods for each modality, simplifying the preprocessing pipeline and ensuring consistency across different types of input.
It is important to note that for images, which contain a large amount of pixel information, the patch version of the model is more suitable. The patch-based approach breaks down images into smaller, manageable segments, allowing the model to process the high-density pixel data more efficiently.

For this paper, which zeroes in on tasks brimming with numerical substance, we meticulously assessed the model's efficacy in tackling assignments steeped in voluminous numerical datasets. Moreover, we opted for octal (base-8) byte sequences as a medium for representing and manipulating data, which offers a structured approach. In addition, the utilization of hexadecimal (base-16) representation is not out of the question; it introduces a degree of adaptability in the encoding and processing of data.

 The mainstream Byte Pair Encoding(BPE) \cite{sennrich2016neuralmachinetranslationrare} can introduce ambiguity and inconsistency in tokenizing numbers, leading to different segmentations of the same number based on context. For instance, the number 12345 might be tokenized as `12', `34', and `5' in one context, or as `1', `23', and `45' in another, losing the inherent meaning of the original numeric value. Additionally, BPE results in fragmented token IDs for numerical entities, such as `7' and `8' being assigned token IDs 4779 and 5014, respectively. This discontinuity in token IDs complicates the management and processing of numerical data, especially when sequential or patterned token IDs are beneficial. Similarly, single-digit tokenization 
 \cite{dubey2024llama}, while straightforward, also leads to discontinuous token IDs for multi-digit numbers, such as 15 being broken down into separate tokens `1' and `5',  each of which is then mapped to an independent token ID. This fragmentation can disrupt the continuity of numerical information, potentially making it more challenging for the model to capture the inherent structure and relationships within multi-digit numbers.

To overcome these limitations, we have crafted an innovative tokenization approach that leverages the binary representation used in computer storage. For textual data, we use UTF-8 encoding to transform characters into byte sequences. When it comes to numerical data, we offer a dual strategy: one involves treating numbers as strings and encoding them in UTF-8, while the other involves converting numbers to their native numerical types before transforming them into byte sequences. This adaptable strategy guarantees that the model can adeptly manage an extensive array of data types, upholding uniformity and efficiency throughout the tokenization process. This approach offers several advantages:
\begin{itemize}
    \item Preservation of Numerical Meaning: By converting numbers into byte sequences, the inherent meaning of the original numeric value is preserved. 
    \item Consistent Tokenization: The byte tokenization method ensures that the same number is always tokenized in the same way, regardless of context, avoiding the inconsistencies seen in BPE.
    \item Continuous Token IDs: Byte tokenization results in a continuous and systematic mapping of numerical values to token IDs. By employing specific computer encoding techniques, numbers like 7 and 8 can be converted into sequential token IDs, thereby streamlining the management and processing of numerical data.
\end{itemize}

Our Byte Tokenization method operates on the principle of encoding all input data into byte arrays, which serve as a common ground for further processing. The general workflow consists of four main steps:
\begin{enumerate}
    \item Data Collection and Preparation: Gather the raw data, including text, numbers, and scientific formulas.
    \item Byte Array Conversion:
\begin{itemize}
    \item Text: Characters are translated into their corresponding Unicode byte values.
    \item Numbers: Numbers can be processed in two distinct manners. First, a number can be treated as a string and subsequently encoded using UTF-8. For instance, the number 12345 would be represented as the string `12345' and then encoded into a byte array[49, 50, 51, 52, 53]. This method is particularly useful for preserving the exact format and any leading zeros that might be significant. Second, a number can be converted into its numerical form (e.g., an integer) and then transformed into a byte array. The integer 12345 can be directly converted into a byte array representation[0, 0, 0, 0, 11, 17, 20, 21]. This approach is more efficient for arithmetic operations and when the numerical value itself is crucial.
    \item Scientific Formulas or Symbols: Complex expressions are parsed and serialized into byte sequences that capture both the structure and content of the formula. In the instance of the formula $E = mc^2$, it would be encoded as a byte array $[69,61,109,99,94,50]$ representing the structure and variables.
  \item Image:Images are typically represented as pixel values, where each pixel can have one or more bytes depending on the color depth. For grayscale images, each pixel is represented by a single byte (8-bit), while for color images (e.g., RGB), each pixel is represented by three bytes (24-bit). The conversion process involves reading the image file and extracting the pixel values into a byte array. Consider a small 2x2 grayscale image, its pixel values, represented as $[128, 64, 192, 255]$, would correspond to the byte array $[128, 64, 192, 255]$ upon conversion.
\end{itemize}
    \item Tokenization: Concatenate the byte arrays from different data types into a single sequence, formatted as follows:
\begin{equation}
\text{bos}\{ \text{text bytearray} \}\{ \text{num bytearray} \}\{ \text{formula bytearray} \}\{ \text{img bytearray} \}\text{eos}
\end{equation}
Here, bos and eos denote the beginning and end of the sequence, respectively, facilitating the identification of boundaries within the processed data.
   \item{Byte Patch Formation}: The byte sequences are systematically divided into fixed-length byte patches based on the Patch Size parameter. In the context of Patch Version tokenization, setting the Patch Size to 16 bytes leads to the formation of each patch containing exactly 16 bytes. For Non-Patch Version tokenization, setting the Patch Size to 1 byte results in the formation of each byte as a standalone token.
\end{enumerate}


\subsubsection{Training }
\label{Training}

\begin{table*}
\centering
\caption{ The input variables for JoI.}
\label{tab:features}
\begin{tabular*}{\textwidth}{@{\extracolsep{\fill}}cc@{}}
\hline
Variable                    &   Definition              \\
\hline
$\rm \Delta$$\eta$  &   difference in pseudorapidity between the particle and the jet axis     \\
$\rm \Delta$$\phi$   &   difference in azimuthal angle between the particle and the jet axis        \\
\hline
$\rm log P_t$                   &   logarithm of the particle's $P_t$      \\
$\rm log E$                     &   logarithm of the particle's energy      \\
$\rm log\frac{P_t}{P_t(jet)}$   &   logarithm of the particle's $P_t$ relative to the jet $P_t$       \\
$\rm log\frac{E}{E(jet)}$       &   logarithm of the particle's energy relative to the jet energy       \\
$\rm \Delta$$R$ &   angular separation between the particle  and the jet axis       \\
$d_0$                          &   transverse impact parameter of the track\\
$d_0$err                       &   uncertainty associated with the measurement of the $d_0$\\
$z_0$                          &   longitudinal impact parameter of the track\\
$z_0$err                       &   uncertainty associated with the measurement of the $z_0$\\
charge                      &   electric charge of the particle     \\
\hline
isElectron                  &   whether the particle is an electron      \\
isMuon                      &   whether the particle is a muon       \\
isChargedKaon             &   whether the particle is a charged Kaon      \\
isChargedPion             &   whether the particle is a charged Pion      \\
isProton             &   whether the particle is a proton      \\
isNeutralHadron             &   whether the particle is a neutral hadron       \\
isPhoton                    &   whether the particle is a photon       \\
\hline
\end{tabular*}
\end{table*}

\textbf{Data Preparation}
A jet consists of several particles, each described by a set of attribute values (e.g., energy, momentum, charge). The attributes of each particle within a jet are listed in Table~\ref{tab:features}, along with the corresponding attribute values presented in the same order.





\begin{figure}[htbp]
    \centering
    \includegraphics[width=.8\textwidth]{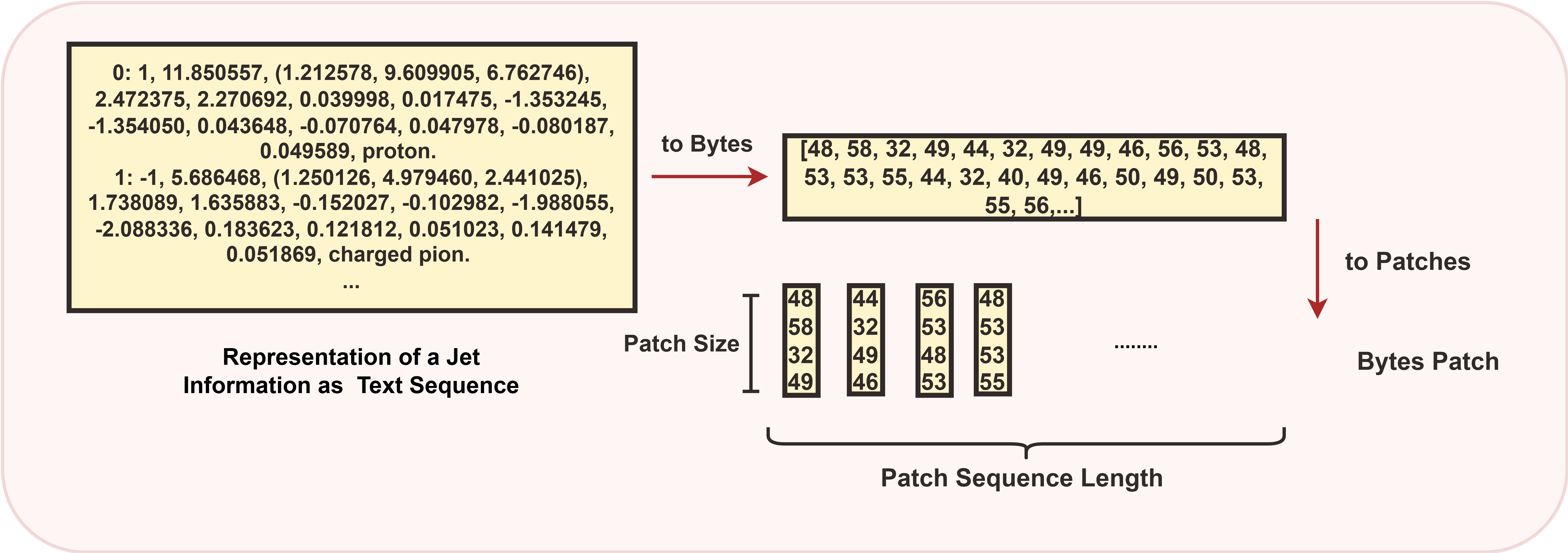}
    \caption{Byte Tokenization:how a jet is converted into a series of byte tokens.
\label{fig:process}}
\end{figure}
In accordance with the tokenization process outlined in Section~\ref{tokenization}, the data undergoes a series of preprocessing steps to transform jet dataset into bytes patch suitable for model training.
We can present a simplified Figure~\ref{fig:process} to illustrate how a jet is converted into a series of bytes patch. To represent jet information in a simple and effective manner, we perform a uniform formatting process on the raw data. In this process, each particle is denoted by an index number, followed by a colon that indicates the attribute values of the particle. These attribute values are arranged in the same order across different particles. These attribute values of different particles are uniformly converted into byte sequences. The byte sequences are then segmented into byte patches of a fixed length defined by the Patch Size, resulting in a total of Patch Sequence Length byte patches. Each byte patch has an input dimension equal to the patch size, and these patches form the input bytes for the model.

Data preprocessing can be implemented in either streaming or non-streaming mode, depending on the dataset size and available memory. 
In non-streaming mode, all data is loaded and processed at once, making this approach suitable for smaller datasets or when sufficient memory is available.
Streaming mode processes data in batches as it is read from the source, which is ideal for large datasets or limited memory scenarios. 
This method optimizes memory usage and processing efficiency.

\textbf{Training Objective}
The JoI task is ideally suited for classification models as it entails assigning input data to specific categories. This classification task requires the model to accurately identify and differentiate between various types of jets based on their features and attributes. Since the model need to determine what type of jet this is,the label are `B-jet', `B-bar-jet', `C-jet', `C-bar-jet', `S-jet', `S-bar-jet', `U-jet', `U-bar-jet', `D-jet', `D-bar-jet' and `G-jet'.

\textbf{Training Details}
During the training process, the model uses the cross-entropy loss function to minimize the difference between the predicted distribution and the true label distribution. The optimizer employed is AdamW, a combination of momentum and adaptive learning rates, which helps improve the training speed and stability of the model. The model has approximately 160 million parameters, and the training is conducted with a batch size of 512 for 30 epochs. The initial learning rate is set to $1 \times 10^{-4}$, and a cosine annealing learning rate schedule is used to adjust the learning rate over the course of training.

Future experiments will explore the performance of models with different parameter scales to further optimize the model's efficiency and effectiveness.
\subsubsection{Experimental Setup}
The experimental framework for this study encompasses the following hardware and software configurations:
\begin{itemize}
\item NVIDIA H100
\item CUDA Version: 12.1
\item DeepSpeed Version: 0.14.4
\item PyTorch Version: 2.3.0
\item Transformers Library Version: 4.43.3
\end{itemize}

\subsection{ParticleNet and Particle Transformer}
\label{PNParT}

ParticleNet \cite{Qu:2019gqs} is based on the EdgeConv operation, which treats a jet as a graph with particles as vertices and connects each particle to its k nearest neighbors to form edges. An MLP is applied to these edges three times, updating vertex features with edge information, allowing information exchange between vertices and edges. The architecture includes three EdgeConv blocks, a global average pooling block, and two fully connected layers with a softmax function for jet predictions.

The Particle Transformer \cite{Qu:2022mxj} uses an encoder-only design with a projection layer for classification scores. It takes particle features and interaction terms between particles as inputs, both of which are embedded using an MLP. The particle embeddings go through L attention blocks with multi-head self-attention, while interaction embeddings enhance the attention mechanism. A class token collects information from all particles via attention and is then passed through an MLP and softmax to produce classification scores.


\section{Conclusion}\label{notesummary}

We presented BBT-Neutron, a task-agnostic model architecture specifically designed for scientific computing tasks, particularly those associated with Big Science experiment infrastructures. Central to this architecture is the Binary Tokenization method, which enables effective training on large-scale numerical data while unifying the pretraining of multiple modalities, including text, numerical data, and images. BBT-Neutron achieves state-of-the-art performance in Jet Origin Identification, a critical task in the data analysis of particle collision experiments. Through comparative analysis, we evaluated the scaling behavior of BBT-Neutron against specialized models Particle Transformer and ParticleNet. The emergent phenomena observed in BBT-Neutron, which are absent in Particle Transformer and ParticleNet, suggest that incorporating domain-specific structures in specialized models may impede the performance phase transitions typically seen in model scaling. We open source BBT-Neutron for the scientific community to build on our work in more areas and tasks. 

\bmhead{Acknowledgements}

We’d like to thank Huiling Qu for his support and valuable feedback throughout the research process. His insights and suggestions have been instrumental in improving the quality of this work. Additionally, we thank Dr.Qi Lu for engaging in fruitful discussions regarding tokenization methods, which have contributed significantly to our approach.


\bibliography{sn-bibliography}

\end{document}